\documentclass[conference]{IEEEtran}
\IEEEoverridecommandlockouts
\usepackage[utf8]{inputenc}

\ifCLASSINFOpdf
\else
\fi
\usepackage{amsmath}
\usepackage{mathtools}
\usepackage{amssymb}
\usepackage{breqn}

\usepackage{color,soul}

\usepackage{url}

\renewcommand{\baselinestretch}{0.92}

\begin{document}
%
\title{Deep Learning for Power System Security Assessment}



\author{\IEEEauthorblockN{Jos\'{e}-Mar\'{i}a Hidalgo Arteaga}
\IEEEauthorblockA{Center of Electric Power and Energy\\
Technical University of Denmark\\
s171800@student.dtu.dk}
\and
\IEEEauthorblockN{Fiodar Hancharou}
\IEEEauthorblockA{Skolkovo Institute of Technology\\
Skolkovo, Russia\\
fiodar.hancharou@skoltech.ru}
\and
\IEEEauthorblockN{Florian Thams and Spyros Chatzivasileiadis}
\IEEEauthorblockA{Center of Electric Power and Energy\\
Technical University of Denmark\\
\{fltha, spchatz\}@elektro.dtu.dk}
\thanks{\emph{This paper has been accepted at IEEE Powertech 2019.}}}


%


\maketitle

\begin{abstract}
Security assessment is among the most fundamental functions of power system operator. The sheer complexity of power systems exceeding a few buses, however, makes it an extremely computationally demanding task. The emergence of deep learning methods that are able to handle immense amounts of data, and infer valuable information appears as a promising alternative. This paper has two main contributions. First, inspired by the remarkable performance of convolutional neural networks for image processing, we represent for the first time power system snapshots as 2-dimensional images, thus taking advantage of the wide range of deep learning methods available for image processing. Second, we train deep neural networks on a large database for the NESTA 162-bus system to assess both N-1 security and small-signal stability. We find that our approach is over 255 times faster than a standard small-signal stability assessment, and it can correctly determine unsafe points with over 99\% accuracy.
\end{abstract}


%
\IEEEpeerreviewmaketitle

\section{Introduction}
Power system security assessment belongs to the most fundamental functions of every system operator. Its aim is to screen a wide range of possible operating points in order to identify a safe operating region and eliminate the possibility of a blackout. Power system operators run different types of security assessment at regular time intervals, ranging from intra-day to every year, checking each operating point against a defined set of instability types, including steady-state stability (e.g. \mbox{N-1} security criterion), transient stability, small-signal stability, and voltage stability \cite{Stability_IEEE}. 

Assessing power system security for every possible operating point is an extremely computationally demanding task. For systems exceeding the size of a few buses, the problem becomes intractable due to the millions of possible operating points. Several approaches have been proposed in the literature, and some of them have also been applied in real power systems, to address this challenge, either through the definition of stability indices that are fast to calculate, through analytical reformulations and approximations (e.g. Lyapunov function \cite{Vu_spchatz}) or different computationally efficient methods \cite{Kundur_PowerSecAssess, VV98}.

Machine learning for transient stability assessment has been applied for the first time in Ref. \cite{Wehenkel_DTbook}, in the form of decision trees. Since then, different approaches have been proposed for different problems, e.g. security assessment \cite{Jensen_NN}, controlled islanding \cite{IREP_2010_1}, and others. With the recent very successful application of deep convolutional neural networks for image and speech recognition, and extremely difficult combinatorial problems (e.g. Go game), deep learning methods appear to offer a promising toolbox with significant potential for power system applications. Assessing power system security by deep learning, however, remains highly unexplored. Ref. \cite{Panciatici_IREP2017} uses machine learning to assess power system security and devise remedial actions. Ref. \cite{DeepL} uses deep learning for extracting valuable features in order to build security rules that can distinguish between safe and unsafe points.

This paper has two main contributions. First, taking advantage of the wide range of deep learning methods for image processing, we introduce for the first time appropriate 2-D representations of power system snapshots as images. Second, using a large database of operating points for the NESTA 162-bus system, we train different deep neural networks for N-1 security \emph{and} small-signal stability assessment, and investigate their performance. Finally, we also briefly discuss ways to integrate such deep neural networks in a security-constrained optimal power flow framework.

This paper is structured as follows. Section~\ref{sec:methodology} presents the methodology we developed to represent power system snapshots and the structure of the neural network we created. Section~\ref{sec:results} presents the results of our case studies on the NESTA 162-bus test system. Section~\ref{sec:opf} discusses possible extensions. Section~\ref{sec:conclusions} concludes.
\section{Methodology}
\label{sec:methodology}
The deep learning algorithm we introduce is based on convolutional neural networks (CNN), which have been successful for applications in image recognition. Inspired by that, in this paper we develop a representation of power system snapshots by images, which we can then feed directly to the convolutional neural network. Considering the rapid development of powerful deep learning algorithms for image processing, an appropriate representation of power system snapshots as images can take advantage of all the state-of-the-art methods developed in the artificial intelligence community. In the following sections, we first present the input data that we used for the training and testing of the neural network, and then we continue with the methodology to create the appropriate images, and the structure of the CNN. 
\subsection{Training and Testing Database}
Our input data are generated for the NESTA 162-bus system \cite{Coffrin2014}. In our previous work, we have introduced an efficient database generation method, which manages to generate hundreds of thousands of datapoints around the security boundary of the system about 10-20 times faster than other state-of-the-art methods \cite{Eff_database}. Using this algorithm, we assess the \mbox{N-1} security and small signal stability of a large dataset of operating points (about 1,000,000) for the NESTA 162-bus test system \cite{Coffrin2014}. 

Each operating point in the database is represented by bus voltages, angles, net active and reactive power demand at each node and active and reactive power in each line. 
The security assessment is based on small-signal stability analysis for a subset of the possible N-1 contingencies $\Bar{C}=[C_1,C_2,...,C_n] $ in the NESTA 162-bus system. 
An operating point is considered safe if the damping ratio $\zeta$ for all contingencies $\bar{C}$ is higher than 3\%, i.e. if $ \min(\zeta(\bar{C}))\geq 0.03 $. 
\subsection{Image Generation}
The images we generate shall (uniquely) represent the state of a power system. To do that, we need to associate the variables extracted from power flows with specific locations in multidimensional arrays. In image recognition algorithms, the images are turned into 2D arrays filled with numerical values that the convolutional neural network can understand. Each colour image is usually converted to three arrays or channels, typically \textit{RGB}, that represent the intensity of red, green and blue colour in each pixel. Following this method, the images created from power system data are also divided in three channels \textit{PQV} that represent the active power, reactive power, and voltage in each snapshot of the power system.

The 2D arrays associated with channels \textit{P} and \textit{Q} will be $N\times N$ matrices, where \textit{N} is the number of buses in the 162-bus case. The diagonal of each matrix has the net active $P_i$ and reactive $Q_i$ power demands at each bus $i$ respectively. If bus \textit{i} and bus \textit{j} are connected by a line, the active $P_{ij}$ and reactive $Q_{ij}$ power flows are placed in the position $(i,j)$ of the matrix while $P_{ji}$ and $Q_{ji}$ are placed in $(j,i)$. Since power losses in the lines are also taken into account, these matrices are not symmetrical. The data from each channel is previously normalised by calculating the absolute value of each element and then dividing by the maximum value of \textit{P} or \textit{Q} in the whole database. All normalised values are between 0 and 1.

\begin{figure}[!t]
\centering
\includegraphics[width=0.73\columnwidth]{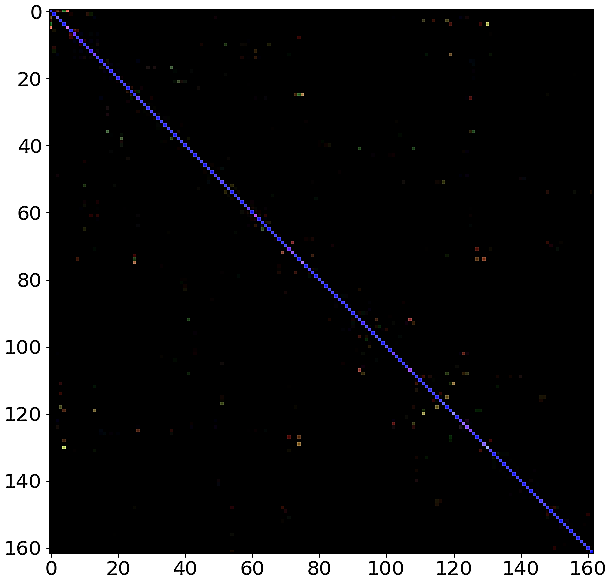}
\caption{Example of image representing a snapshot of NESTA 162-bus power system}
\label{fig:milkyway}
\vspace{-0.4cm}
\end{figure}

The 2D array associated with channel \textit{V} has also size $N\times N$, and its diagonal is filled with the voltages $V_i$ in each bus. The off-diagonal elements, $i\neq j$, are filled with the absolute values of the voltage drops $V_{ij}$ along each line. Unlike the other two channels, matrices in channel $V$ are symmetrical. The data in this channel (bus voltage and line voltage drop) are normalised as well, similarly to the other channels. In the three channels \textit{PQV}, the matrices are sparse because the power system has only 284 lines for the 162 buses. An example of an image can be seen on Figure \ref{fig:milkyway}, where the three channels \textit{PQV} are superimposed and plotted as if they were \textit{RGB}, creating a coloured picture. The coloured pixels are always placed in the same spots but with different intensities. Even though all snapshots are different, two different snapshots are usually visually similar with a naked eye -- but not for a computer. In his Nobel lecture in 2011 \cite{Universe}, Saul Perlmutter explained that an early way of searching supernovae was to take images from distant galaxies periodically and subtract the negative from an older image to the latest image. Both images would look the same but when there is a supernova, the subtraction leaves a single white point over a dark background. If this same procedure is followed with our created images, it will be possible to spot the differences in all snapshots.
An image $X_s\in\mathbb{R}^{N,N,c}$ created from a snapshot $s$ of the power system is a 3D tensor\footnote{The term ``tensor'' refers to multidimensional arrays or vectors, used in TensorFlow (\url{https://www.tensorflow.org/}), the machine learning library we use. We will use the terms ``tensor'' and ``array'' interchangeably.} with size $[N,N,c]$, and $c$ being the number of channels. 
The total input data set $X\in\mathbb{R}^{S,N,N,c}$ with all snapshots turned into images becomes a 4D tensor with $[S,N,N,c]$ size, where $S$ is the number of snapshots that the database contains. Due to the large size of $X$ (104.8 GB), only small batches, $X_b\in\mathbb{R}^{b,N,N,c}$ where $b$ is the batch-size and $b<<S$, are generated when needed in the training process.

The output data, a vector containing zeros and ones depending on the security state of each snapshot, is turned into labels by one-hot encoding having as a result an output data set $y\in\mathbb{R}^{S,N_c}$ where $N_c$ is the number of labels: in our case two (safe/unsafe). In \cite{one-hot}, one-hot encoding is explained as a method to encode categorical variables having $n$ categories as $n$-dimensional feature vectors. For each category, one of the positions of the vector is filled by a one and the rest by zeros, resulting in a vector space where each category is orthogonal and equidistant to the rest. \emph{Unsafe} snapshots are now represented as [1, 0] while \emph{safe} snapshots as [0, 1].

Once the input $X$ and output $y$ sets are constructed, all the data is split into training $(X_{train} \textup{ and } y_{train})$, validation $(X_{val} \textup{ and } y_{val})$ and test sets $(X_{test} \textup{ and } y_{test})$. The training set contains 70\% of the total set while the validation and test sets contain the remaining 10 \% and 20\% of the data respectively. Initially, the dataset is shuffled to assure that all subsets have the same share of safe cases (about 14.5\%). As $X$ does not fit in the memory of the computers used to train the CNN, it is impossible to create the full size tensor $X$ before training. Instead, we split the number of snapshots S in training, validation and test sets, and generate smaller batches $X_b$ that we use as input to the CNN.

\subsection{Structure of the Deep Neural Network}
\begin{figure}[h]
\centering
\includegraphics[width=\columnwidth]{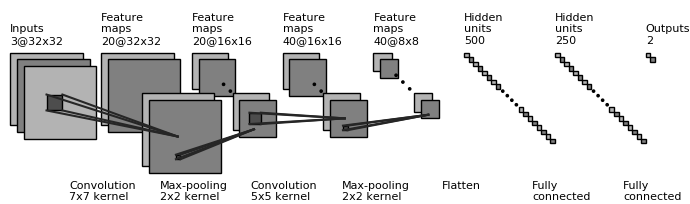}
\caption{Structure of a typical CNN. For the exact structure of the CNN developed for this paper see Table~\ref{CNN}.}
\label{CNNflow}
\end{figure}

Convolutional neural networks (CNN) have been designed for processing data in the form of multiple arrays. As explained above, our model has to process three 2D arrays that represent $P$, $Q$ and $V$ in the studied power system. The structure of a convolutional net has several stages. The initial stages are composed by a series of convolutional and pooling layers. The output of these initial stages is reshaped in a flattening layer and then fed into fully-connected layers. The typical structure of a CNN is shown in Fig.~\ref{CNNflow}.
The whole convolutional net works with the same principle as a multi-layer perceptron (MLP). Equation \eqref{MLP}, where $g$ is the activation function and $W_{ij}$ and $b_j$ are the weights and biases from node $i$ to node $j$ respectively, shows how the output of each node is calculated:
\begin{equation}\label{MLP}
    X_{j} = g(\sum_{i=1}^{F}(W_{ij}X_i+b_{j}))
\end{equation}
The different layers that form the CNN, as well as its training process, are explained in the following paragraphs.
\subsubsection{Convolutional Layer}
Convolutional layers take as input 2D arrays and train a series of 2D kernels or filters that aim to recognise patterns in the input arrays giving the local weighted sum as an output. 

The input to any convolutional layer is a 4D batch $X_b$ with $[b,H,W,F_{i-1}]$ size that is turned into a $[b,H,W,F_{i}]$ tensor where $H$ and $W$ are the height and width of the 2D arrays respectively and $F_{i}$ and $F_{i-1}$ are the number of filters or nodes in layer $i$ and in the previous one respectively.

The weights in a convolutional layer are a 4D tensor with a $[K,K,F_{i-1},F_{i}]$ size, with $K$ being the kernel size, while the biases are a 1D tensor of $F_{i}$ size. The kernel size and the number of filters are tuning parameters chosen by the user.

The most commonly used activation function in convolutional layers is the Rectified Linear Unit (ReLU) \cite{relu}, whose function is given by $f(x)=max(0,x)$ \cite{Lecun_ConvNets_IEEEProc}. 

\subsubsection{Max-Pooling Layer}
The pooling layers do not have weights and biases associated and their aim is to reduce the spatial size of the convolutional layers' output. Max-pooling is the most commonly used type of pooling, and also the one used in our algorithm. Max-pooling layers apply a moving window, in our case of size $2\times 2$, over a 2D input and give the maximum value from every sub-region covered by the window. 

This type of layer is applied after each convolutional layer, giving as an output a $[b,\frac{M}{2},\frac{N}{2},F_{i}]$ size tensor. The activation function is applied after the max-pooling layer, instead of right after the convolutional layer, because it is less costly due to its smaller sized input.

\subsubsection{Flattening Layer}
Flattening layers connect convolutional or pooling layers to fully-connected layers. This type of layer reshapes the output of the last max-pooling layer into a 1D array per snapshot in order to be able to feed the first fully-connected layer. The size of the output of this layer $[b,M\cdot N\cdot F_i]$ is the result of compacting the three dimensions from the previous layer $[b,M,N,F_i]$. 

\subsubsection{Fully-connected Layer} here, each node is connected to all the nodes from the previous layer. The output of each layer is a $F_{i}$-size vector per snapshot. The first fully-connected layer is fed by the flattening layer. The weights in these layers are a 2D tensor with a $[F_{i-1},F_{i}]$ size while the biases are a $F_{i}$-size vector. In all fully-connected layers except the last one, 20\% of the nodes are disconnected while training in order to avoid over-fitting (i.e. 20\% dropout rate).

All fully-connected layers except for the output layer are activated by a ReLU function. The output layer, which is the one that finally performs the classification, is activated by a Softmax function or normalised exponential function \cite{soft}. Softmax equations are shown in \eqref{softmax} and \eqref{softmax2}, where $\sigma$ represents the output, $N_c$ is the number of classes and $X_i$ is the input vector. The Softmax function turns a $N_c$-dimensional vector of real values into a vector of the same size with values between 0 and 1 where all elements sum 1. This function is used in the output layer in order to get the probability distribution over the different classes, in our case safe and unsafe. The CNN chooses the class with the highest probability to occur as the predicted security state. 
\begin{align}
    & \sigma : \mathbb{R}^{N_c} \rightarrow \lbrace \sigma\in\mathbb{R}^{N_c} \mid \sigma_i>0,\sum_{i=1}^{N_c}\sigma_i=1\rbrace \label{softmax}\\
    & \sigma(X)_i = \frac{e^{X_i}}{\sum_{i=1}^{N_c}e^{X_i}} \quad for \quad i=1,...,N_c \label{softmax2}
\end{align}
\subsection{Training the Neural Network}
In order to measure the classification performance of our neural network during training, we need to define a loss function. The loss function is based on the cross entropy between the estimated and real output probability distributions. The cross entropy equation across $m$ snapshots is shown in \eqref{loss} as the first term, where $y_i$ and $\hat{y}_i$ are the real and predicted probabilities of class $i$ respectively. The first term of the cross entropy, which corresponds to the positive class predictions, is multiplied by a class coefficient $\Phi$ in order to penalise more false positive predictions ($\Phi > 1$) or false negative predictions ($\Phi < 1$). The coefficient $\Phi$ is useful when the labels of the data are unbalanced. Based on the loss function presented in \cite{DeepL}, the estimated recall $\hat{Rec}$, specificity $\hat{Spe}$, precision $\hat{Pre}$ and F1-score $\hat{F1}$, whose definitions are explained later, are added multiplied by the negative costs $\alpha_r$, $\alpha_s$, $\alpha_p$ and $\alpha_f$ respectively. These costs are negative because we aim to maximise the algorithm's performance on those metrics. The total loss function includes also L2 regularisation at the end in order to avoid over-fitting and also penalise large values of weights.
\begin{dmath}\label{loss}
    Loss = -\frac{1}{m}\sum_{i=1}^{m}\big[\Phi y_i log(\hat{y}_i)+(1-y_i) log(1-\hat{y}_i)\big]-\alpha_r\hat{Rec}-\alpha_s\hat{Spe}-\alpha_p\hat{Pre}-\alpha_f\hat{F1}+\frac{\lambda}{2}\sum_{j=1}^{N_L}W_j^2
\end{dmath}
We use the Adam optimiser \cite{Adamopt}, which is an algorithm based on stochastic gradient descent, to minimise the loss function while tuning the trainable parameters of the neural network: weights and biases. The optimiser is initialised with an initial learning rate chosen by the user. The CNN is trained by applying a mini-batch gradient descent method, i.e. the parameters are updated every time a batch, smaller than the training set, goes through the optimiser. These small batches are created and overwritten right after their use in order to be able to fit this learning process in the available memory space. The whole training set, and therefore all the batches, go through the optimiser in every epoch. After every epoch, the model is validated in the validation set. The final model will be the one with the highest validation accuracy in all the epochs. Before training, a large number of epochs is chosen but in order to avoid over-fitting, early stopping is applied with a patience parameter chosen by the user. This means that if the validation loss is not decreasing over the epochs, the training stops and the saved weights and biases are the ones corresponding to the epoch with the highest validation accuracy.

\subsection{Proposed Model}
The proposed CNN architecture initially has three consecutive series of convolutional and max-pooling layers. The kernel size of the layers is $9\times9$, $7\times7$ and  $5\times5$ for the first, second and third convolutional layers while the number of filters is 20, 40 and 80 respectively. The kernel size decreases from the first layer because the input to the convolutional layers gets smaller. This allows to increase the number of filters in the next layers.

\begin{figure}[h]
\centering
\includegraphics[width=0.8\columnwidth]{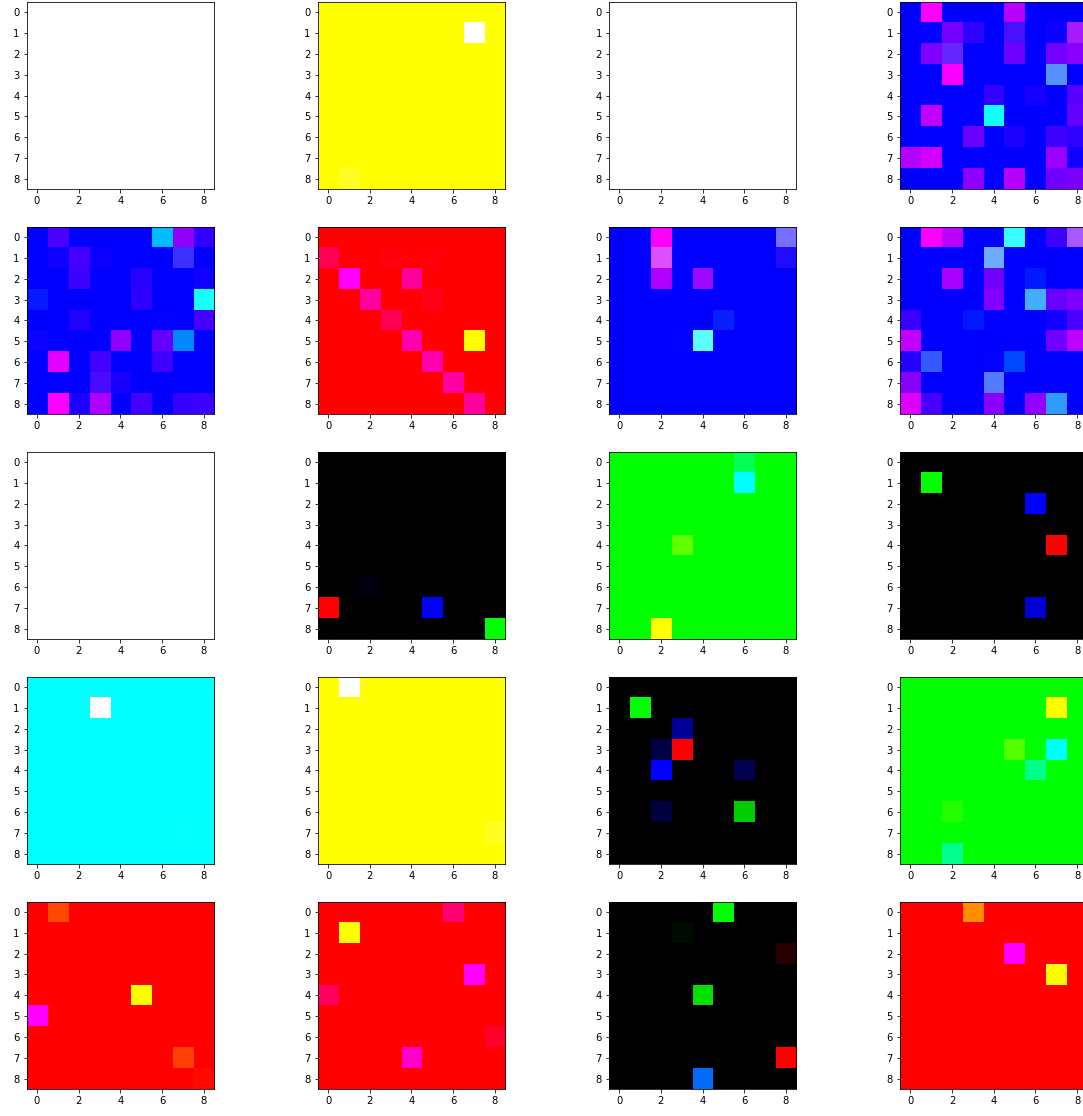}
\caption{Image representing the weights in Conv1.}
\label{wc1}
\vspace{-0.3cm}
\end{figure}

Figure \ref{wc1} shows the weights corresponding to the first convolutional layer. This image has been created by normalising the weights and superimposing the three input channels \textit{PQV} into one as if they represented \textit{RGB} channels. There is twenty $9\times9$ images due to the number of filters and their size in the first convolutional layer. Red, green and blue colours represent large weights in channels \textit{P}, \textit{Q} and \textit{V} respectively and small in the rest. Black colour represents small weights in the three channels while white colour shows large weights in all of them. This colourful image can only be obtained for the first layer since in the following ones, the inputs are more than the initial three.
The flattening layer is applied after the last max-pooling layer. Then, the first fully-connected layer, with 250 nodes is fed. This layer feeds directly to the output layer which has 2 nodes, one per label. Table \ref{CNN} shows the size of each layer as well as the shapes of the weights and biases. The number of parameters to train in each layer is also represented. The number and size of the filters as well as the number of nodes have been chosen heuristically. 
\begin{table}[h]
\renewcommand{\arraystretch}{1}
\caption{Representation of CNN architecture.}
\label{CNN}
\centering
\begin{tabular}{|c|c|c|c|c|}
\hline
\textbf{Layer} & \textbf{Shape}     & \textbf{Weights size} & \textbf{Bias size} & \textbf{\# param}  \\ \hline
Input                 & {[}b,162,162,3{]}  &                        &                     &                    \\ \hline
Conv1                 & {[}b,162,162,20{]} & {[}9,9,3,20{]}         & {[}20{]}            & 4.880              \\ \hline
Max-pool1             & {[}b,81,81,20{]}   &                        &                     &                    \\ \hline
Conv2                 & {[}b,81,81,40{]}   & {[}7,7,20,40{]}        & {[}40{]}            & 39.240             \\ \hline
Max-pool2             & {[}b,40,40,40{]}   &                        &                     &                    \\ \hline
Conv3                 & {[}b,40,40,80{]}   & {[}5,5,40,80{]}        & {[}80{]}            & 80.080             \\ \hline
Max-pool3             & {[}b,20,20,80{]}   &                        &                     &                    \\ \hline
Flatten               & {[}b,32000{]}      &                        &                     &                    \\ \hline
FC1                   & {[}b,250{]}        & {[}32000,250{]}        & {[}250{]}           & 8.000.250          \\ \hline
FC2 (output)          & {[}b,2{]}          & {[}250,2{]}            & {[}2{]}             & 502                \\ \hline
\textbf{Total}        &                    &                        &                     & \textbf{8.124.952} \\ \hline
\end{tabular}
\end{table}
The optimiser is fed with a batch-size of 128 snapshots, and a maximum of 200 epochs with a patience of 30 epochs. The initial learning rate is set to 0.001. 

The model has been implemented using the Python open source machine learning library TensorFlow for GPU and trained in a single Tesla V100 GPU core from DTU's HPC cluster for 23 hours (DTU cluster limits the time use of each job to a maximum of 24 hours). For our simulations, this time was enough for the validation error to converge to a minimum, even though there were always some small fluctuations in the error during the training process. It is believed that training the CNN in several GPU cores in parallel would speed up the process and improve the results further.
\section{Results}
\label{sec:results}
In this section, the results obtained by the presented model are shown. First, we describe the evaluation metrics used to measure how good our model is. Subsequently, we present the scores obtained by testing the model, and at the end we show the substantial reduction of computation time by using the presented model over traditional methods.
\subsection{Evaluation Criteria and Model Performance}
The evaluation criteria used are based on the confusion matrix, shown in Table \ref{ConfusionMatrix}. The confusion matrix is filled with the True Positives (TP), referring to the number of snapshots predicted as unsafe that are actually unsafe, True Negatives (TN), i.e., the number of snapshots predicted as safe that are actually safe, and False Positives (FP) and False Negatives (FN) as the snapshots that have been predicted as the opposite class from what they actually are.
\begin{table}[h]
\caption{Confusion Matrix.}
\label{ConfusionMatrix}
\centering
\begin{tabular}{c|c|c|}
\cline{2-3}
\textbf{}                                    & \textbf{Predicted Unsafe} & \textbf{Predicted Safe} \\ \hline
\multicolumn{1}{|c|}{\textbf{Actual Unsafe}}   & TP                      & FN                        \\ \hline
\multicolumn{1}{|c|}{\textbf{Actual Safe}} & FP                      & TN                        \\ \hline
\end{tabular}
\end{table}
The following measures are extracted from the different elements in the confusion matrix. 
\begin{itemize}
    \item \textbf{Recall}: Also named as Sensitivity or True Positive Rate (TPR), it is the proportion of correct positive predictions in all the positive cases. Calculated as $Recall = \frac{TP}{TP+FN}$.
    \item \textbf{Specificity}: Also called True Negative Rate (TNR), it is the proportion of correct negative predictions in all the negative cases. Calculated as $Specificity = \frac{TN}{TN+FP}$.
    \item \textbf{Precision}: Also known as Positive Predictive Value (PPV), is the proportion of correct positive predictions in all positive predictions. Calculated as $Precision = \frac{TP}{TP+FP}$.
    \item \textbf{F1-Score}: The harmonic mean of \textit{Precision} and \textit{Recall}. Calculated as $F1 = 2\frac{Precision \cdot Recall}{Precision+Recall}$.
    \item \textbf{Accuracy}: The proportion of correct predictions in all data set. Calculated as $Accuracy = \frac{TP+TN}{TP+FP+TN+FN}$.
    \item \textbf{Matthews correlation coefficient}: A measure of correlation between predicted and actual values in a binary classification. It yields a coefficient between -1 and 1, with 1 being when the prediction is perfect, 0 when the prediction is no better than random, and -1 when all predicted values are mistaken. Calculated as $MCC = \frac{(TP\cdot TN)-(FP\cdot FN)}{\sqrt{(TP+FP)(TP+FN)(TN+FP)(TN+FN)}}$.
\end{itemize}
The measure of accuracy is often enough to measure the quality of the classification for a database with the same number of safe and unsafe operating points. Since this is not the case in our problem, where only 14.56\% of the cases are safe, other measures become also meaningful when evaluating the classification. It also needs to be mentioned that it is substantially more important to predict correctly unsafe cases than safe cases. In our case, a FN would have a much higher negative impact (and cost) for the power system than a FP. For that reason, it is more important to have a high \textit{Recall} than a high \textit{Specificity}. 

\begin{figure}[h]
\centering
\includegraphics[width=\columnwidth]{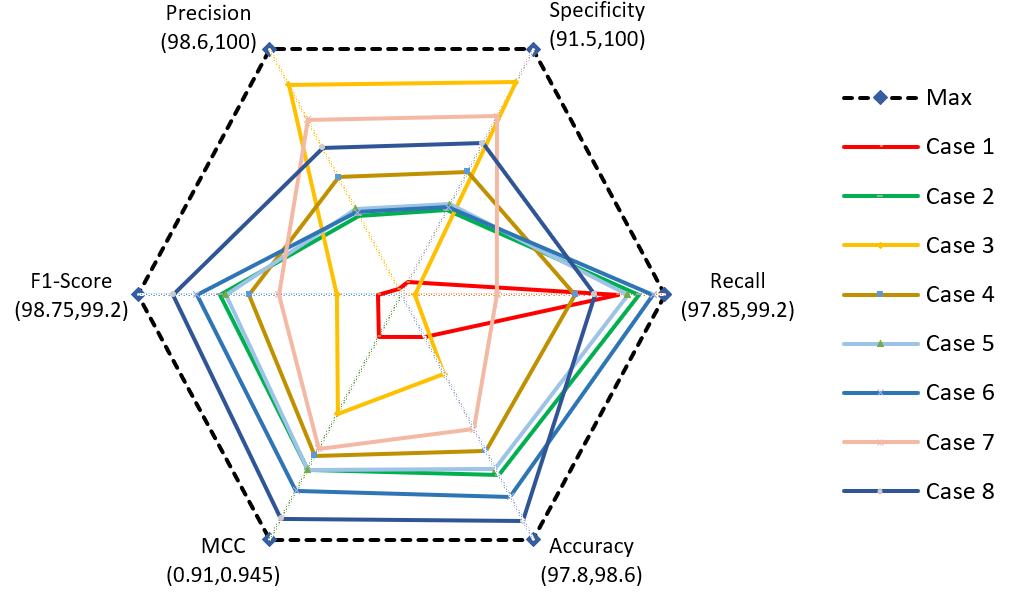}
\caption{Radar chart of the scores in the different cases. Table~\ref{Cases} describes the used parameters for each case. Note that each axis has a slightly different range in order to enhance the readability of the figure.}
\label{radar}
\end{figure}

As it has just been mentioned, the database labels are unbalanced, containing a substantially higher number of unsafe cases compared with safe cases. If we were looking for balancing the cross entropy loss, the balanced class coefficient $\Phi_{b}$ would be $\Phi_{b}=\frac{S_{safe}}{S_{unsafe}}=0.17$ where $S_{safe}$ and $S_{unsafe}$ are the number of safe and unsafe snapshots respectively. Since we aim to minimise the FNs because predicting correctly the unsafe cases is more important, the class coefficient has to be $\Phi>\Phi_{b}$; in our studied cases, as shown in Table~\ref{Cases} it was $\Phi\geq1$. The metric costs vector is defined as $\Bar{\alpha}=[\alpha_r,\alpha_s,\alpha_p,\alpha_f]$ (see \eqref{loss}). The metrics directly related to predicting unsafe cases correctly and therefore decreasing FNs are \textit{Recall}, \textit{Precision} and \textit{F1-Score}, while \textit{Specificity} is related to the safe cases prediction.
\begin{table}[h]
\caption{Cases depending on $\Phi$ and $\Bar{\alpha}$.}
\label{Cases}
\centering
\begin{tabular}{|c|c|c|}
\hline
\textbf{Cases} & \textbf{$\Phi$} & \textbf{$\Bar{\alpha}=[\alpha_r,\alpha_s,\alpha_p,\alpha_f]$} \\ \hline
\textbf{1}     & 1               & {[}0, 0, 0, 0{]}                                              \\ \hline
\textbf{2}     & 2               & {[}0, 0, 0, 0{]}                                              \\ \hline
\textbf{3}     & 5               & {[}0, 0, 0, 0{]}                                              \\ \hline
\textbf{4}     & 1               & {[}0.5, 0.5, 0.5, 0.5{]}                                      \\ \hline
\textbf{5}     & 1               & {[}0.5, 0, 0.5, 0.5{]}                                        \\ \hline
\textbf{6}     & 2               & {[}0.5, 0, 0.5, 0.5{]}                                        \\ \hline
\textbf{7}     & 3               & {[}0.5, 0, 0.5, 0.5{]}                                        \\ \hline
\textbf{8}     & 2               & {[}0, 0, 0.5, 0.5{]}                                          \\ \hline
\end{tabular}
\end{table}
Table \ref{Cases} shows the different studied cases, with different values of $\Phi$ and $\Bar{\alpha}$, and Figure \ref{radar} shows the best scores of our model in the test set in each case. The radar chart aims to demonstrate how the variation of $\Phi$ and $\Bar{\alpha}$ can influence the training outcome. Note that the scale of each axis in this diagram is specified under each label.

Cases 1-3 show the variation of $\Phi$ while $\Bar{\alpha}$ is filled with zeros. It can be observed that a value close to the optimal $\Phi$ could be around 2 since increasing it more is counterproductive. In case 4, all the values from $\Bar{\alpha}$ are increased while $\Phi$ stays 1 and the results are improved in comparison with case 1. The addition of evaluation metrics in the loss function results in an improvement of performance. Increasing $\Phi$ and $\Bar{\alpha}$ too much can result in worse scores due to the trade-off between the importance of the different terms in the loss function. Since having a high \textit{Specificity} is less important than the rest, cases 5-7 have $\alpha_s$ as zero while the rest of elements in $\Bar{\alpha}$ stay 0.5. It can be observed again that when $\Phi$ equals 2, in case 6, there is a peak of performance. In case 6 we obtain the highest score in \textit{Recall} with a 99.14\% meaning that from all the combinations tried, the one that predicts the best the unsafe states is case 6. Investigating the confidence interval, we found that with 99\% likelihood the score in \textit{Recall} in case 6 will belong to the confidence interval 99.14\% $\pm$ 0.092\%. On the other hand, in case 8 where both $\alpha_r$ and $\alpha_s$ are zero, the model scores the best in \textit{F1-Score}, MCC and accuracy with a 99.14\%, 0.942 and 98.54\% respectively. Again, the 99\% confidence intervals of these scores in case 8 are \textit{F1-Score}: 99.14\% $\pm$ 0.092\%, \textit{MCC}: 0.942 $\pm$ 0.00097, and \textit{Accuracy}: 98.54\% $\pm$ 0.12\%. In this case, the overall classification performs better than in the rest of cases but the correct classification of unsafe classes decreases in exchange of a rise in MCC and accuracy. 

In order to identify a pattern in the miss-classified operating points in case 8, the whole operating area covered by the database is split in clusters by \textit{k-means} clustering \cite{5453745}. This technique is applied several times for a number of 3, 5, 10 and 20 clusters. Once the operating space is divided in clusters, the clusters where the miss-classified points belong are identified. This gives as a result that at least the 99\% of the miss-classified points belong always to the same cluster. This means that most of these points belong to a small specific region of the security boundary that the CNN failed to model.
\subsection{Computation Time}
The main goal of applying deep learning techniques to power system security assessment is to devise algorithms that provide results much faster than the standard small-signal stability analysis methods for all possible contingencies. For that reason, we measure the time needed to perform a security assessment with the presented model and compare it to the time spent in computing all necessary small-signal stability analyses. To ensure a fair comparison, (i) both methods are tested in the same machine: a single core from an Intel Xeon Processor E5-2650 v4 of the DTU HPC cluster, and (ii) the computation time reported is the \emph{average} time after performing the security assessment on a single snapshot for 1000 times. Table \ref{Time} reports the average time per test over these 1000 tests.
\begin{table}[h]
\caption{Time tests: Small-signal stability analysis vs. ConvNet model.}
\label{Time}
\begin{tabular}{c|c|c|}
\cline{2-3}
                                           & \textbf{Small-signal stability analysis} & \textbf{ConvNet model} \\ \hline
\multicolumn{1}{|c|}{Average Time } &                  21'950~ms                        &   86~ms              \\ \hline
\end{tabular}
\end{table}
As it can be observed in Table \ref{Time}, the time spent to perform a security assessment for the studied power system is reduced about 255 times, making the presented model way more efficient for security assessment than traditional methods.

\section{Discussion and possible extensions}
\label{sec:opf}
The used training database (online available, see \cite{Eff_database}) contains all possible N-1 contingencies for all possible operating points for a given demand profile (over 1,000,000 points). Topology changes in the form of N-1 investigations are already considered. Future work shall consider a wider range of topology changes and uncertainty in demand.

Inspired by our previous work, where we have reformulated Decision Trees for stability assessment to a Mixed Integer Linear Program \cite{ThamsHalilbasic_IREP2017} or Mixed Integer Second Order Cone Program \cite{Halilbasic_PSCC_2018} in order to fully consider N-1 security and small signal stability in Optimal Power Flow, in future work we will also examine extensions of our deep learning algorithm to AC-OPF formulations. Through an iterative framework, the CNN can help determine the binding constraints that lead to a N-1 secure and small-signal stable point, which can then be inserted in the optimization problem. In our preliminary investigations, the average computation time of such a setup after carrying out 50 tests in an Intel Core i7-8550-U CPU is 19.99s. Future work will focus on computation efficiency and convergence guarantees.
\section{Conclusions}
\label{sec:conclusions}
This paper has presented a method based on convolutional neural networks and deep learning for power system security assessment. There are two main contributions. First, we represent for the first time power system snapshots as images that can be easily processed by convolutional neural networks. Acknowledging the fact that safe or unsafe operating points exhibit similar patterns, we represent power flows between buses as 2-dimensional images. Taking advantage of the wide range of deep learning methods available for image processing, we can achieve a remarkable performance of deep neural networks for power system security assessment. Second, we develop and extensively test different deep neural networks to assess N-1 security \emph{and} small-signal stability on the NESTA 162-bus system. We find that our approach is over 255 times faster than a standard small-signal stability assessment for a single operating point, while it achieves an accuracy of over 98\% and a Recall of over 99.14\% (i.e. classifying unsafe states as unsafe). As discussed in the last section, future work will focus on the integration of deep learning for power system security assessment to an optimal power flow framework.
\section*{Acknowledgment}
\footnotesize{This work has been partially supported by the multiDC project, funded by Innovation Fund Denmark, Grant Agreement No. 6154-00020B.}
\bibliographystyle{IEEEtran}
\bibliography{diss_biblio}
\end{document}